\documentclass[letterpaper]{article}

\usepackage{natbib,alifeconf}  
\usepackage{xcolor}
\usepackage{amsmath}
\usepackage{todonotes}
\usepackage{listings}
\usepackage{protobuf/lang}  
\usepackage{protobuf/style}
\usepackage{hyperref}
\usepackage[font=tiny]{subfig}
\usepackage[font=scriptsize,labelfont=bf]{caption}
\usepackage{float}

\DeclareGraphicsExtensions{.png,.pdf}
\definecolor{seafoam}{RGB}{62, 175, 118}

\title{Growing 3D Artefacts and Functional Machines with Neural Cellular Automata}

\author{Shyam Sudhakaran$^1$, Djordje Grbic$^1$, Siyan Li$^1$, Adam Katona$^2$\\ \Large{Elias Najarro$^1$, Claire Glanois$^3$,  Sebastian Risi$^1$}\\
\mbox{}\\
$^1$ IT University of Copenhagen, $^2$ University of York, $^3$ Shanghai University\\
shyamsnair@protonmail.com, djgr@itu.dk, lisiyansylvia@gmail.com, ak1774@york.ac.uk, sebr@itu.dk
} 

\begin{document}
\maketitle

\begin{abstract}
  Neural Cellular Automata (NCAs) have been proven effective in simulating morphogenetic processes, the continuous construction of complex structures from very few starting cells. Recent developments in NCAs lie in the 2D domain, namely reconstructing target images from a single pixel or infinitely growing 2D textures. In this work, we propose an extension of NCAs to 3D, utilizing 3D convolutions in the proposed neural network architecture. Minecraft is selected as the environment for our automaton since it allows the generation of both static structures and moving machines. 
We show that despite their simplicity, NCAs are capable of growing complex entities such as castles, apartment blocks, and trees, some of which are composed of over 3,000 blocks. Additionally, when trained for regeneration, the system is able to regrow parts of simple functional machines, significantly expanding the capabilities of simulated morphogenetic systems. The code for the experiment in this paper can be found at: \url{https://github.com/real-itu/3d-artefacts-nca}. 
\end{abstract}

\section{Introduction}
Life begins as an embryo. From the single cell, the interacting forces of morphogenesis and cell growth then create tissues and organs, stacking them together into a functional organism. Morphogenesis, the shape organization of cell populations, has been a focus in regenerative medicine for tissue engineering and organ assembly \citep{joshi2012epithelial,kinney2014engineering,nakashima2003application,hubbell2003materials}. These cell populations operate on a local level, communicating information to immediate neighbors. Despite relying solely on local dependency, cells are able to form complex biological entities, consisting of various organs and the connections between them. This local dependency also allows for very useful properties, such as growth and regeneration from a small set of cells.


One approach for understanding morphogenes is via emulating local interactions of cells in \textit{Cellular Automata} (CA). CAs contain grids of cells iteratively updated based on cell-level rules. The specific update of a cell depends upon the states of the cell itself and the neighboring cells. Therefore, these rules can be expressed as a function of a cell's and its neighbors' states. In Conway's Game of Life \citep{games1970fantastic}, a less complex case, this function only counts the neighboring alive and dead cells.
When equations alone are unable to encapsulate the desired update rules, Neural Networks are often introduced to replace the function. Models resulting from such substitutions are referred to as \textit{Neural Cellular Automata} (NCA).

The sophisticated nature of morphogenesis makes NCAs a promising modeling approach. Works on simulating morphogenesis with NCAs have primarily focused on easily-replicable artefacts, such as pixelated emojis \citep{mordvintsev2020growing} and 2D images \citep{ruiz2021neural}. Recent progress has also demonstrated its capability in generating high-fidelity 3D shapes \citep{kim2021learn3d} and regenerating damaged morphology \citep{horibe2021regenerating}. However, significant gaps still exist between the current state-of-the-art and real-life applications for morphogenesis simulation.

\begin{figure*}[!htb]
  \centering
  \label{ref_label1}\includegraphics[width=0.85\textwidth]{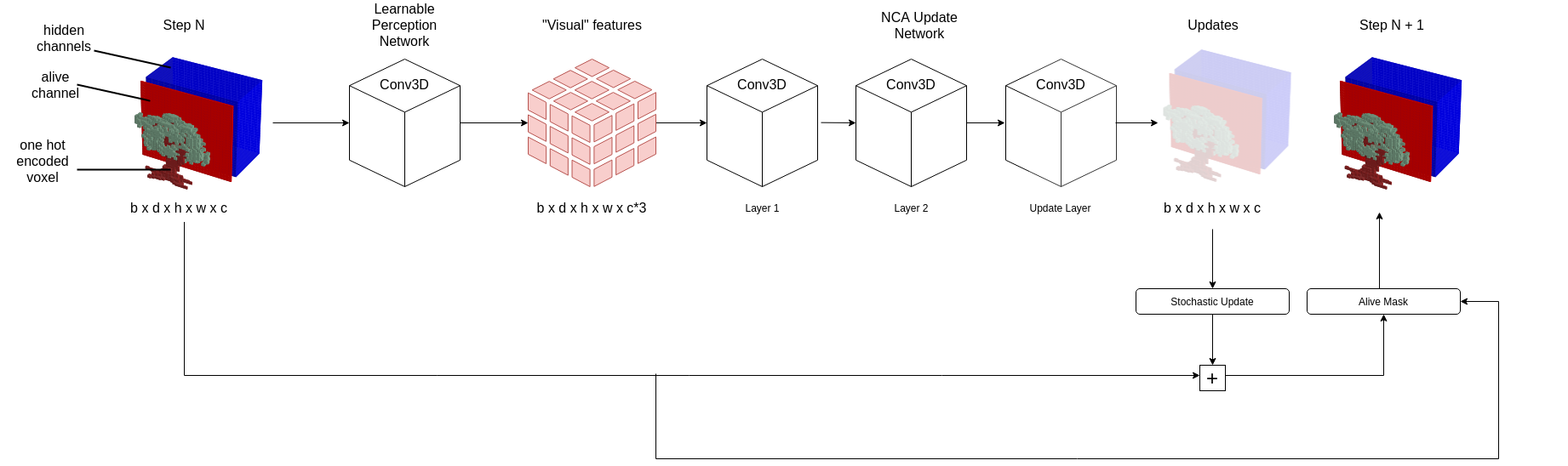}
  \caption{\label{nca_architecture} 3D Neural Cellular Automata update rule.
  }
  \label{fig:overview}
\end{figure*}

To improve the generalizability of NCAs to real-world applications, 
we propose an extension of \cite{mordvintsev2020growing} to facilitate complex structure generation in three dimensions. Specifically, we develop a 3D NCA for generating 3D Minecraft structures and functional robots (Figure~\ref{fig:overview}).

We train out NCA to grow complex structures with up to 3,584 blocks and 50 unique block types through a series of reconstruction tasks on Minecraft designs. 
Minecraft is a good test domain for our approach because it allows the creation of both static (e.g.\ an oak tree) and dynamic (e.g.\ a simple flying machine) structures. The results show that, in most cases, NCAs scale to 3D effectively, able to generate complex entities from a single cell. The NCAs also hold the same "regenerative" properties as their counterparts in 2D.

The main contributions of our work include: (1) An extension of NCA in 3D on multi-class voxels; (2) A cellular automation for generating Minecraft structures of varying complexity, as well as moving functional robots. In addition, we develop a particular loss function that should be generally helpful in domains that require an NCA to deal with varying discrete block types. 

\section{Related work}

\subsection{Cellular Automata}
Originally proposed in 1940s~\citep{neumann1966theory}, Cellular Automata mimic developmental processes in multi-cell organisms including morphogenesis and irregular growth. The architecture contains a grid of similarly structured cells, which are updated periodically in discrete time steps. At every time step, the status of each cell can be represented as a state, which is then transitioned into the next state per the update rule. The specific transition depends on the current state of the cell and the neighboring cells. Despite individual cells being simplistic, the entire CA is capable of complex behaviors. Cellular Automata have been speculated to aid understanding of biological pattern formations \citep{wolfram1984cellular}, modeling of lattice-based physical systems in percolation and nucleation \citep{VICHNIAC198496}, and synthesis of artificial life \citep{LANGTON1986120}. 

\subsection{Neural Cellular Automata}
CAs that replace the update rules with a neural network are referred to as \textit{Neural Cellular Automata} (NCA). \cite{li2002neural} models development of multiple land use utilizing a densely-connected neural network to simulate conversion probabilities between different land types.
\cite{nichele2017neat} replaces the update process of CAs with a continuously evolved \textit{Compositional Pattern Producing Network} (CPPN) to replicate simple 2D patterns; the network maps neighboring cell states to the next state for the central cell. 

The primary inspiration for our work is \cite{mordvintsev2020growing}, where a NCA  learns to generate specific 2D images starting from a single pixel. Every cell within the CA encodes cell state information using a continuous size-16 state vector, representing the RGB values, whether a cell is alive, and hidden channels for simulating local morphogens concentrations. Sobel filters estimate the partial derivatives of cell state channels, which are then concatenated with individual cell state vectors to form the ``perception vector" of the cell. This perception is fed into a series of neural layers to evoke a potential update of the cell. Sharing the neural network layers throughout the CA enforces a universal, differentiable update rule.
Trained via supervised learning, the model can be driven towards a goal using a reconstruction loss, more specifically the error between the target image pixels and those that were generated. The work also includes a ``sample pool", which helps relieve issues with ``catastrophic forgetting". Additional operations include stochastic update of cells to imitate independent development of individual cells, as well as alive-masking to prevent participation of empty (``dead") cells in the update process.

\cite{ruiz2021neural} extends upon \cite{mordvintsev2020growing} by learning NCA's manifold, where each point encode a different update rule corresponding to a different image. The weights of the neural NCA's layers are functions of the target image, after passage through an encoder. Although improving generalizability, the resulting model is still constrained to 2D images. 

\cite{kim2021learn3d} proposes a Generative Neural Cellular Automata for generating diverse 3D structures from an arbitrary starting shape. Instead of pre-defined, the update rules are neural network outputs from a Markov Chain. As \cite{ruiz2021neural}, this enables the model to generate multiple structures, contrary to the typical NCA, which is only specific to one target. One limitation is that there is only one type of building block. They also only predict the "outer" layer of the 3d structures, improving efficiency with the added sparseness, but the tradeoff is the lack of detail within the interior of a structure.

\cite{horibe2021regenerating} equips simulated soft robots with the ability to partially regenerate and regain locomotive capacity using two NCAs, one for generating the initial morphology and the other for restoration once the robot is damaged. The transition for a voxel (cell) is determined by a neural network using cellular states from the neighboring four voxels. A ``voxel cost" is also introduced to prevent box-like morphologies. We employ a similar cost, using an intersect over union cost to steer the model to only generate cells that it needs to reconstruct the target.

Compared to previous literature, our work has higher dimensionality (3D instead of 2D) than \cite{mordvintsev2020growing} and \cite{ruiz2021neural}, more construction unit types than \cite{kim2021learn3d}, and we capture all surrounding cells with a 3D convolution instead of only the four immediate neighbors as in \cite{horibe2021regenerating}.

\begin{table*}[h]
    \centering
    \begin{tabular}{ |p{0.2\textwidth}||p{0.1\textwidth}|p{0.1\textwidth}|p{0.1\textwidth}|p{0.1\textwidth}| p{0.1\textwidth}  }
         \hline
         \multicolumn{5}{|c|}{Minecraft Entity Details} \\
         \hline
         Entity& No. Unique Blocks & No. Non Air Blocks & Padding & Padded Size (W$\times$D$\times$H)\\
         \hline
         Village House   & 10 & 84 & Yes &   10$\times$10$\times$10\\
         Blacksmith & 17 & 280 & Yes & 10$\times$ 10$\times$10\\
         MiniCastle & 35 & 1253 & Yes & 20$\times$20$\times$20\\
         Jungle Temple & 15 & 1283 & Yes & 20$\times$20$\times$20\\
         Tree    & 3 & 1622 & Yes & 30x30x30\\
         Apartment Block & 50 & 3136 & No & 18$\times$18$\times$23\\
         Cathedral & 23 & 3584 & No & 33$\times$27$\times$31\\
         Flying Machine & 6 & 8 & Yes & 10$\times$10$\times$10\\
         Caterpillar& 7  & 137 & No & 8$\times$27$\times$6\\
         \hline
    \end{tabular}
    \caption{Minecraft Entity Details}
    \label{tab:entity_details}
\end{table*}

\begin{table*}[h]
    \centering
    \begin{tabular}{ |p{0.2\textwidth}||p{0.1\textwidth}|p{0.1\textwidth}|p{0.2\textwidth}| p{0.1\textwidth} | p{0.1\textwidth} | p{0.1\textwidth} }
         \hline
         \multicolumn{6}{|c|}{Experiment Hyperparameter Details} \\
         \hline
         Entity & No. hidden & Min \& Max steps & NCA Update Net layer 1 \& 2 no. channels & layer init. stdev & lr \\
         \hline
         Village House   & 10 & (48, 64) & (32, 32) & 0.1 & 0.0002\\
         Blacksmith & 10 & (48, 64) & (32, 32) &  0.1 & 0.002\\
         MiniCastle & 10 & (48, 64) & (32, 32) & 0.1 & 0.002\\
         Jungle Temple & 12 & (48,64) & (64, 64) & 0.1 & 0.002\\
         Tree    & 12 & (64,64) & (64, 64) & 0.1 & 0.002\\
         Apartment Block & 12 & (64, 65) & (64, 64) & 0.1 & 0.002\\
         Cathedral & 12 & (50,51) & (64, 64) & 0.2 & 0.002\\
         Flying Machine & 10 & (48, 64) & (32, 32) & 0.001 & 0.002\\
         Caterpillar & 12 & (48, 64) & (64, 64) & 0.02 & 0.002\\
         \hline
    \end{tabular}
    \caption{Experiment Hyperparameter Details}
    \label{tab:hyperparam_details}
\end{table*}
\section{Approach: Growing Minecraft Entities}

Similar to \cite{mordvintsev2020growing}, we teach a Neural Cellular Automata to generate 3D entities through reconstruction tasks. We represent a Minecraft entity as a 3D grid of cells, each with a cell state vector containing channels for (1) its block type; (2) its living status; (3) its hidden states. The block type is embedded into a one-hot vector, whose entries correspond to unique block types in the structure, including the empty ``air" block as its first entry. We reproduce the mechanism for cells' living status from  \cite{mordvintsev2020growing}: Each cell state has an ``alive channel" with an alpha value; a cell is ``alive" when it or one of its neighbors has an alpha value greater than 0.1 and ``dead" otherwise. ``Dead" cells are nullified by setting the ``air" block channel to 1 and everything else to 0. A cell's hidden states are represented as a continuous vector. This vector is used to carry information through steps and differentiate individual cells.

\subsection{Model Architecture}
As previously stated, we replace the 2D convolutions in \cite{mordvintsev2020growing} with 3D ones to allow generation of 3D structures. Furthermore, we also employ a learnable perception network instead of the static one with Sobel filters. The dynamic perception net is implemented as a 3D convolutional layer with \texttt{kernel size = 3}, \texttt{stride = 1}, and \texttt{output channels = cell\_state\_channels * 3}. \newline The cell updates are retrieved by passing the visual features from the perception net into 2 configurable linear layers, which are implemented as 3D convolutions with \texttt{kernel size = 1} and \texttt{stride = 1} (See configurations: Table \ref{tab:hyperparam_details}, which are passed into a non configurable update layer. Like the previous work in 2D, we use a stochastic update, multiplying half of the updates by zero, as a form of "dropout" (\cite{JMLR:v15:srivastava14a}), which helps with overfitting. We then apply an "alive mask" to the updates, which uses the interaction with the living channel stated before. This is implemented by multiplying the updates with the outputs of a MaxPool layer and a boolean mask where 1 if $>$ 0.1 and 0 otherwise. We initialize all models with standard normal initialization, with mean = 0 and standard deviation ranging from 0.0001 to 0.1

\begin{figure*}[!htb]
  \centering
  \subfloat[Normalized IOU / Structural Loss]{\label{ref_label1}\includegraphics[width=0.8\textwidth]{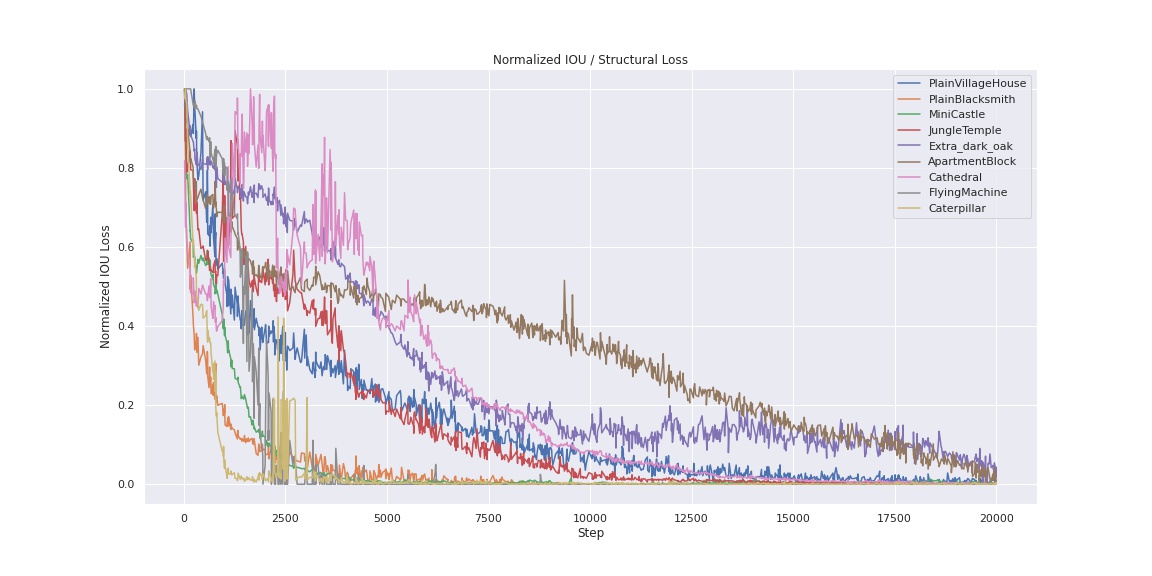}}
  
  \subfloat[Normalized Total Loss]{\label{ref_label2}\includegraphics[width=0.8\textwidth]{ 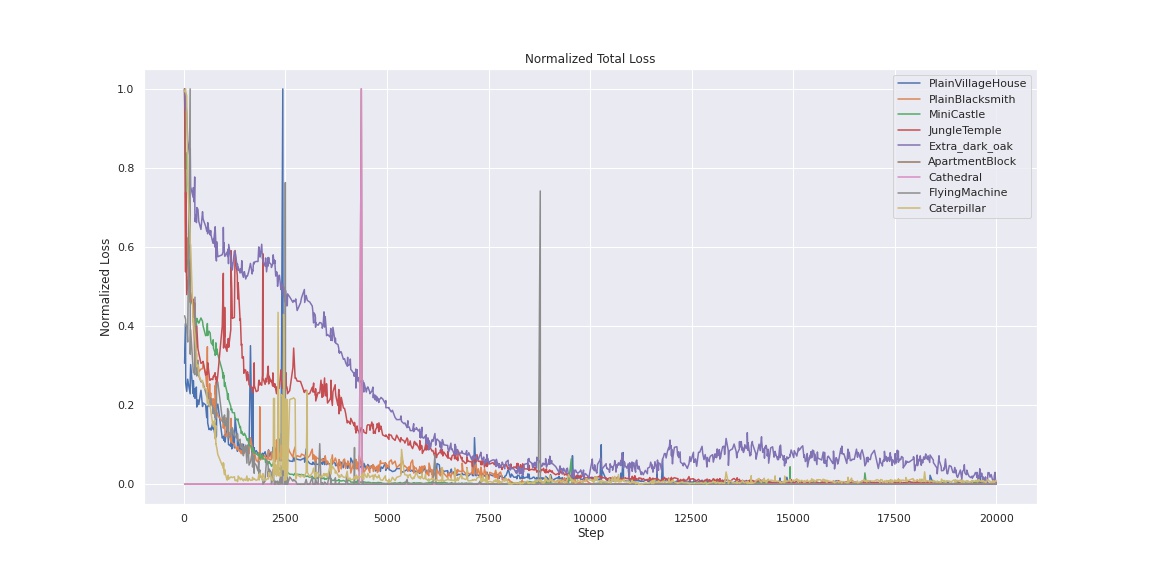}} 
  \caption{\label{training_curves} Normalized training curves for total loss and IOU / Structural loss}
\end{figure*}

\subsection{Training Procedure}
We use a similar training process as the original work where the NCA tries to generate a target entity from a single living cell. optimizing over a reconstruction loss using supervised learning. However, because each cell can only be of a single block type, we treat the structure reconstruction task as a multi-class classification problem, predicting the type of a given cell. Therefore, our objective becomes the minimization of the cross entropy loss between cells from the target structure and from the predictions. We utilize a combination of \textit{LogSoftmax} and \textit{NLLLoss}, provided by Pytorch's \texttt{cross\_entropy} method. With just this loss, we noticed that the training performance was unstable and the model demonstrated a preference for ``air" blocks. We hypothesize this to be due to an imbalance in our training data. ``Air" blocks often take up the majorities of our selected structures, and these blocks can be either alive or dead. Such an imbalance might result in the model over-predicting ``air" blocks. This may also be alleviated by removing padding, but doing so causes training instability for most cases.

To address this, we divide up the loss calculations between cells that are classified as non-``air" blocks and the cells that are classified as ``air" blocks. We also add an \textit{Intersect Over Union (IOU)} cost, which measures the absolute difference between the non-``air" blocks in the target and the generated entities. We anticipate this additional cost component to impose structural constraints on generated structures and improve regeneration accuracy.
\begin{align}
    CE(\hat{y}_{i,c}, y_{i}) &= -I(y_{i} = c)\hat{y}_{i,c} + \log{\sum_{j=1}^M(exp(\hat{y}_{i,j}))}\\
    Inter(\hat{y}, y) &= \sum_{}(I(\operatorname*{argmax}_c{\hat{y}_{c}} > 1) \cap I(y > 1))\\
    Uni(\hat{y}, y) &= \sum_{}(I(\operatorname*{argmax}_c{\hat{y}_{c}} > 1) \cup I(y > 1))
\end{align}
\begin{align}
    IOU(\hat{y}, y) &= (Uni(\hat{y},y) - Inter(\hat{y},y)) / (Uni(\hat{y},y) + 1e^{-8})\\
    Loss &= \frac{1}{N}\sum_{i=1}^N(\sum_{c=1}^M(CE(\hat{y}_{i,c},c)) + IOU(\hat{y}_{i}, y_{i})
\end{align}

When a generated entity's non-``air" blocks are identical to its respective target, the IOU cost is 0. The gradients for this loss function are accumulated over time, with multiple forward passes ranging from anywhere between 48 to 64 steps, see Table \ref{tab:hyperparam_details}.

Consistent with \cite{mordvintsev2020growing}, the system becomes unstable when generating samples for more steps than what it was trained for. To alleviate this, we include a ``sample pool" of size 32, which is updated by the outputs of each batch. The pool is initialized with a set of "seed states", which are composed of  a single living cell. Every training iteration a batch of samples is taken from the pool. The best-performing sample from each batch (the sample that has the `least loss compared to the target) is replaced with a single living cell to prevent catastrophic forgetting. The sample pool imitates experience replay. The outputs of the batch are obtained by passing the batch into the NCA for a random number of forward steps, parameterized by ``min\_steps" and ``max\_steps". At the end of the training iteration, the sample pool is updated with the outputs. 

\subsection{EvoCraft environment}
To interact with Minecraft, we use the  EvoCraft API \citep{grbic2020evocraft}. EvoCraft includes a Python programmable interface that allows reading and writing blocks. The API provides a connection over the local network to a running Minecraft server. The interface contains an enumeration of supported block types and block orientations. 

The three available functions are \emph{spawnBlocks}, \emph{readCube}, and \emph{fillCube}. \emph{spawnBlocks} takes a list of \emph{Block} objects and sends them server side to be spawned in the Minecraft world. A \emph{Block} object contains the block type, block orientation, and a coordinate where the block should be spawned.
Similarly, the \emph{readCube} function takes a bounding box corner coordinates and returns a list of Block objects representing the blocks contained within the box. This function reads the current state of the world with regards to block positions and to follow if the blocks are moving through the world (redstone moving contraptions). Finally, the \emph{fillCube} function takes a bounding box corner coordinates and a block type. The result is the box filled with blocks of that type. 
These three methods allow us to implement a cellular automaton by rearranging the blocks depending on the current state.

The API uses gRPC protocol to send/receive messages and the interface definition (e.g.\ Listing~\ref{listingService}). Most of the block types have no functional purpose except aesthetics, while some have unique functional purposes and interactions with other blocks.  Blocks like “clay”, “terracotta”, “stone”, “glass”, “dirt” are inert but have different textures and colors making them useful to build aesthetically pleasing structures. Blocks like “lava” and “concrete powder” can interact with “water” blocks to produce “obsidian” blocks and “concrete” blocks. Similarly, “lava” and “TNT” blocks can destroy if they come into contact with or explode near to nearby blocks. Most interestingly for our purposes, there are “piston” and “sticky piston” blocks that push/pull other blocks if they are adjacent to a “redstone” block. “Slime” blocks glue to the adjacent blocks and pull them if the “slime” block gets pushed/pulled by a “piston” block. The existence of these blocks allows players to build large and dynamic moving structures. The unique interactions between the blocks provides ways for the Minecraft world to be dynamic independent of the rules of the cellular automaton operating beyond. For the full list of supported blocks consult \citep{grbic2020evocraft} appendix.

\begin{lstlisting}[language=protobuf2, style=protobuf, breaklines=true, caption={The EvoCraft gRPC API definition. For brevity the headers and message definitions are omitted.}, label=listingService, captionpos=b]
service MinecraftService {
    /** Spawn multiple blocks. */
    rpc spawnBlocks (Blocks) returns (Empty)

    /** Return all blocks in a cube */
    rpc readCube (Cube) returns (Blocks)

    /** Fill a cube with a block type */
    rpc fillCube (FillCubeRequest) returns (Empty)
}
\end{lstlisting}

\begin{figure*}[!htb]
  \centering

  \subfloat[Target MiniCastle]{\label{ref_label1}\includegraphics[width=0.15\textwidth]{ 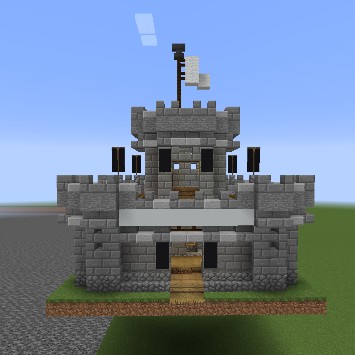}} \hspace{0.05\textwidth}
  \subfloat[MiniCastle step -- 10]{\label{ref_label1}\includegraphics[width=0.15\textwidth]{ 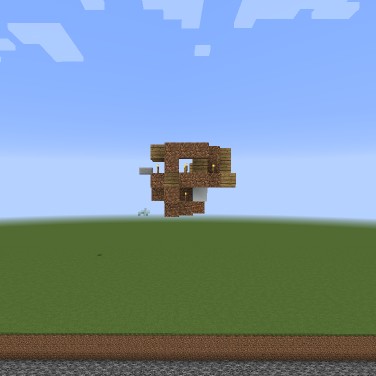}} \hspace{0.01\textwidth}
  \subfloat[MiniCastle step -- 25]{\label{ref_label2}\includegraphics[width=0.15\textwidth]{ 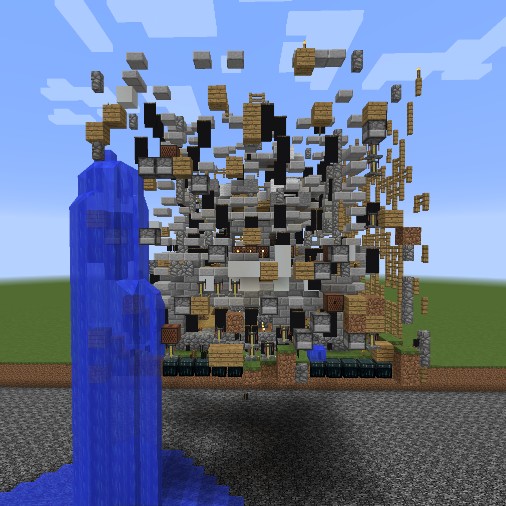}} \hspace{0.01\textwidth}
  \subfloat[MiniCastle step -- 50]{\label{ref_label3}\includegraphics[width=0.15\textwidth]{ 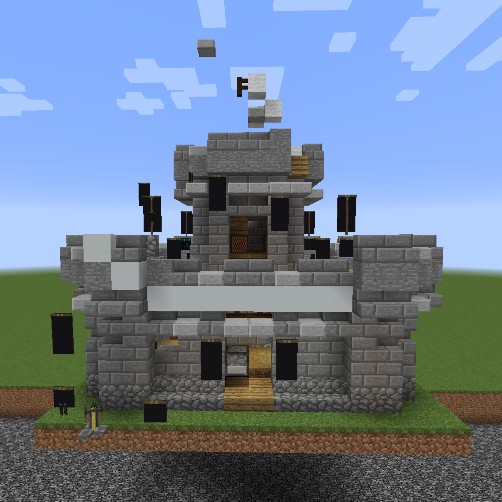}} \hspace{0.01\textwidth}
  \subfloat[MiniCastle step -- 100]{\label{ref_label4}\includegraphics[width=0.15\textwidth]{ 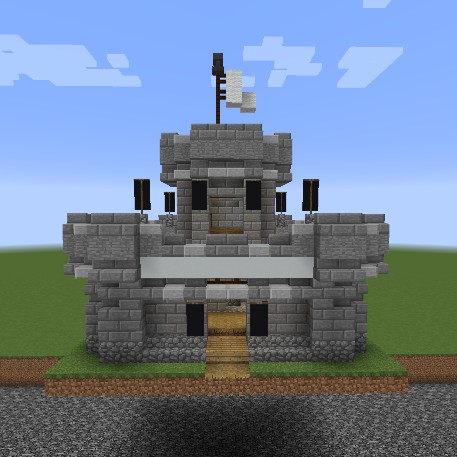}} \hspace{0.01\textwidth} 

  \subfloat[Target JungleTemple]{\label{ref_label3}\includegraphics[width=0.15\textwidth]{ 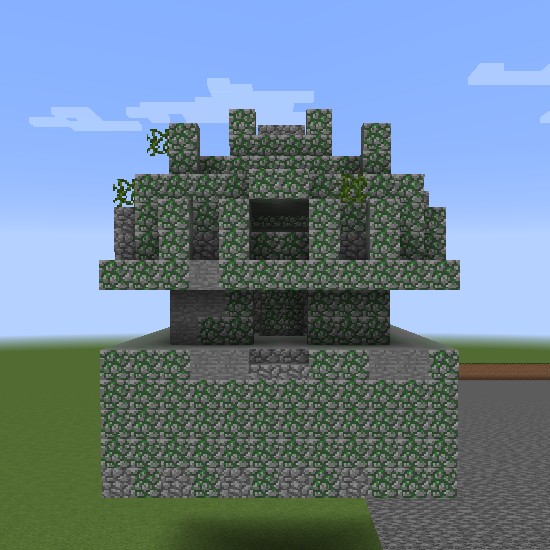}} \hspace{0.05\textwidth}
  \subfloat[JungleTemple step -- 10]{\label{ref_label1}\includegraphics[width=0.15\textwidth]{ 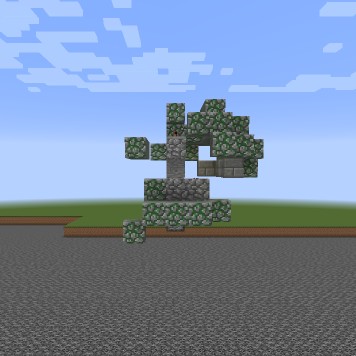}} \hspace{0.01\textwidth}
  \subfloat[JungleTemple step -- 25]{\label{ref_label2}\includegraphics[width=0.15\textwidth]{ 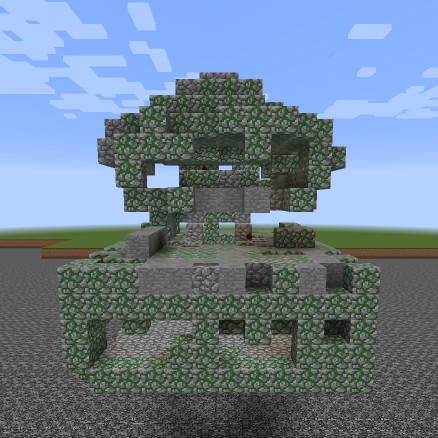}} \hspace{0.01\textwidth}
  \subfloat[JungleTemple step -- 50]{\label{ref_label2}\includegraphics[width=0.15\textwidth]{ 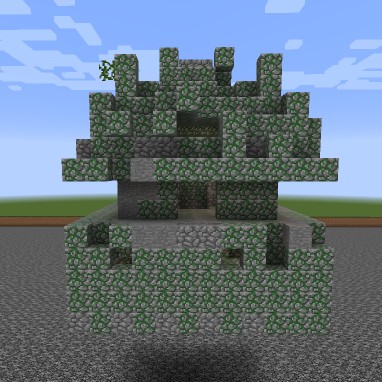}} \hspace{0.01\textwidth}
  \subfloat[JungleTemple step -- 100]{\label{ref_label4}\includegraphics[width=0.15\textwidth]{ 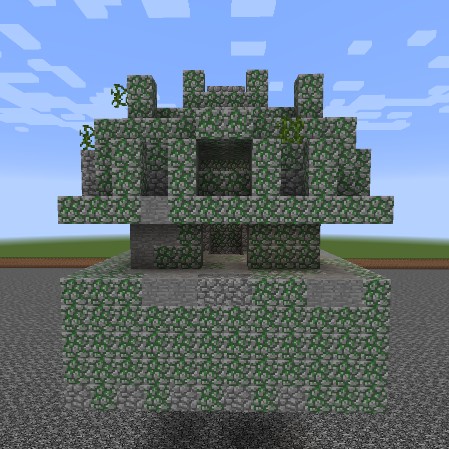}} \hspace{0.01\textwidth}

  \subfloat[Target Tree]{\label{ref_label3}\includegraphics[width=0.15\textwidth]{ 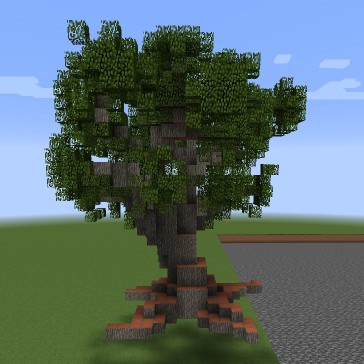}} \hspace{0.05\textwidth}
  \subfloat[Tree step -- 10]{\label{ref_label1}\includegraphics[width=0.15\textwidth]{ 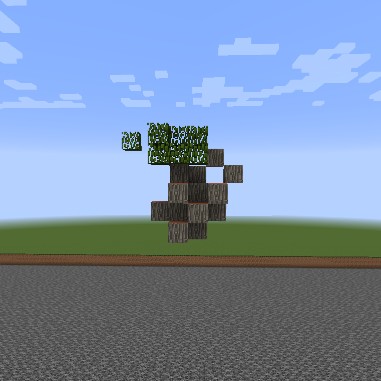}} \hspace{0.01\textwidth}
  \subfloat[Tree step -- 25]{\label{ref_label2}\includegraphics[width=0.15\textwidth]{ 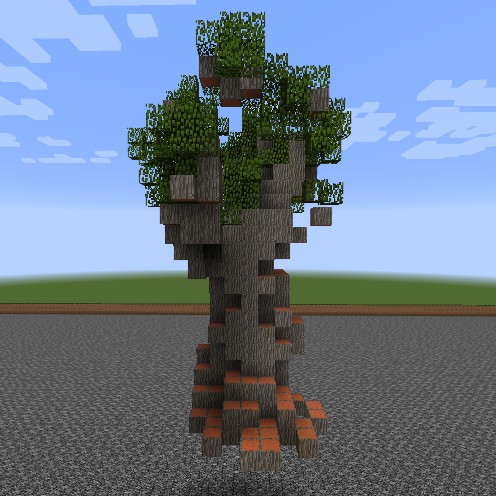}} \hspace{0.01\textwidth}
  \subfloat[Tree step -- 50]{\label{ref_label2}\includegraphics[width=0.15\textwidth]{ 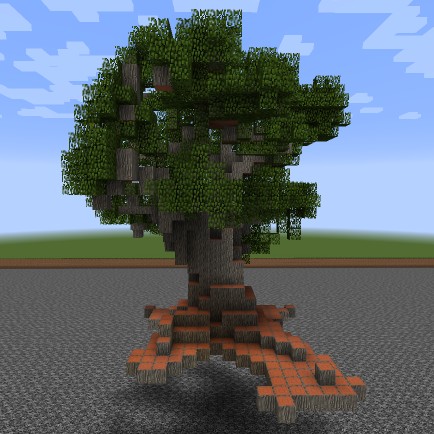}} \hspace{0.01\textwidth}
  \subfloat[Tree step -- 100]{\label{ref_label4}\includegraphics[width=0.15\textwidth]{ 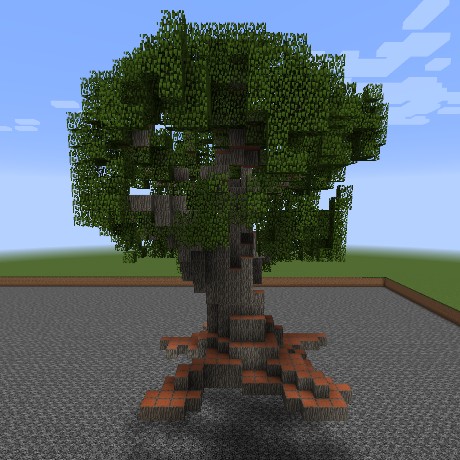}} \hspace{0.01\textwidth}
 
  \subfloat[Target ApartmentBlock]{\label{ref_label3}\includegraphics[width=0.15\textwidth]{ 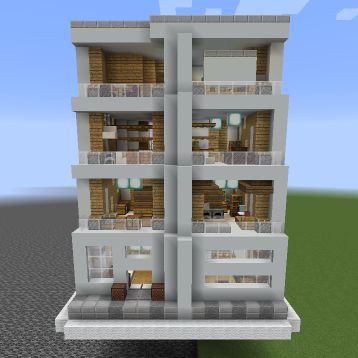}} \hspace{0.05\textwidth}
  \subfloat[ApartmentBlock step -- 10]{\label{ref_label1}\includegraphics[width=0.15\textwidth]{ 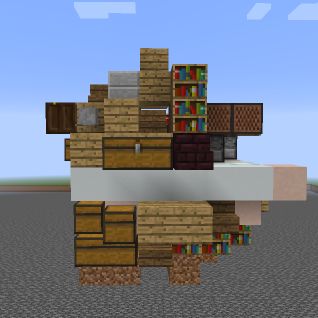}} \hspace{0.01\textwidth}
  \subfloat[ApartmentBlock step -- 25]{\label{ref_label2}\includegraphics[width=0.15\textwidth]{ 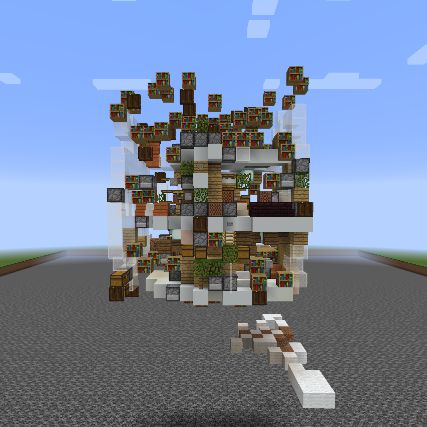}} \hspace{0.01\textwidth}
  \subfloat[ApartmentBlock step -- 50]{\label{ref_label2}\includegraphics[width=0.15\textwidth]{ 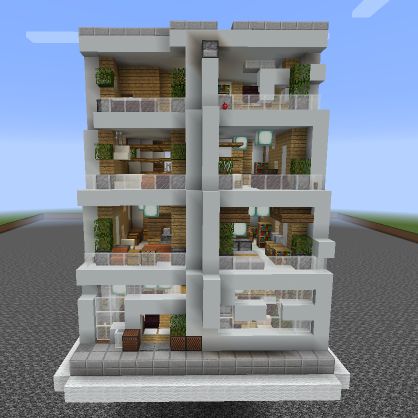}} \hspace{0.01\textwidth}
  \subfloat[ApartmentBlock step -- 100]{\label{ref_label4}\includegraphics[width=0.15\textwidth]{ 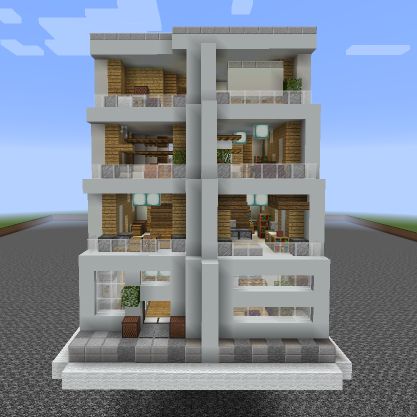}} \hspace{0.01\textwidth}

  \caption{\label{structure_generation} Structural generation over time}
\end{figure*}

\begin{figure*}[!htb]
  \centering
  \subfloat[Target Cathedral]{\label{ref_label3}\includegraphics[width=0.15\textwidth]{ 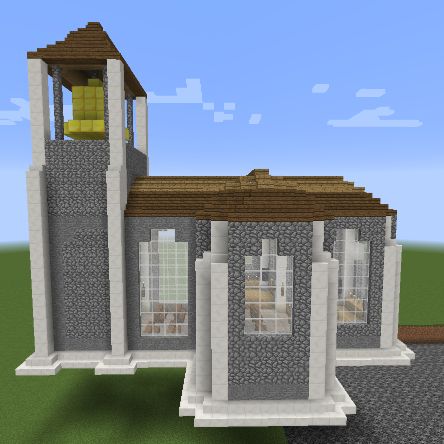}} \hspace{0.05\textwidth}
  \subfloat[Cathedral step -- 10]{\label{ref_label1}\includegraphics[width=0.15\textwidth]{ 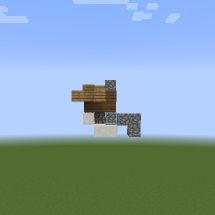}} \hspace{0.01\textwidth}
  \subfloat[Cathedral step -- 25]{\label{ref_label2}\includegraphics[width=0.15\textwidth]{ 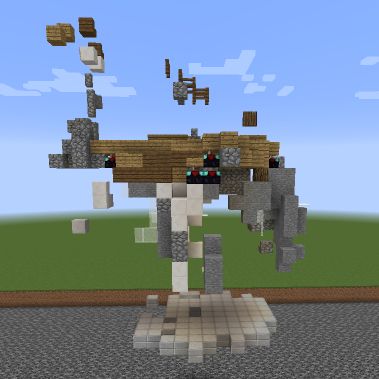}} \hspace{0.01\textwidth}
  \subfloat[Cathedral step -- 50]{\label{ref_label2}\includegraphics[width=0.15\textwidth]{ 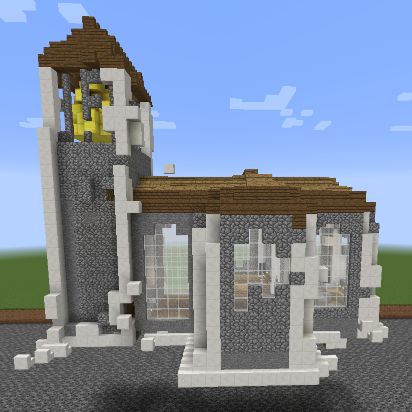}} \hspace{0.01\textwidth}
  \subfloat[Cathedral step -- 100]{\label{ref_label4}\includegraphics[width=0.15\textwidth]{ 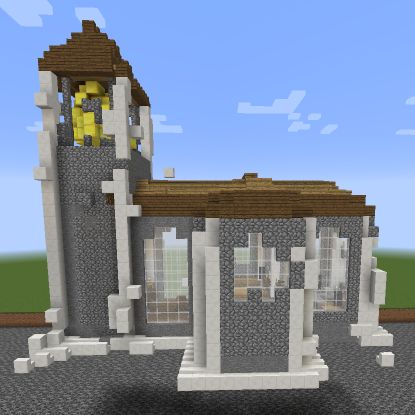}} \hspace{0.01\textwidth}

  \caption{\label{cathedral_structure_generation} Cathedral generation failure case}
\end{figure*}

\subsection{Dataset}
\begin{figure*}[!htb]
  \centering
  \subfloat[Target FlyingMachine]{\label{ref_label1}\includegraphics[width=0.15\textwidth]{ 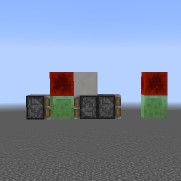}} \hspace{0.05\textwidth}
  \subfloat[FlyingMachine -- step 10]{\label{ref_label1}\includegraphics[width=0.15\textwidth]{ 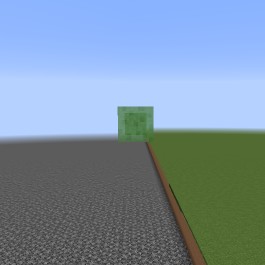}} \hspace{0.01\textwidth}
  \subfloat[FlyingMachine -- step 15]{\label{ref_label2}\includegraphics[width=0.15\textwidth]{ 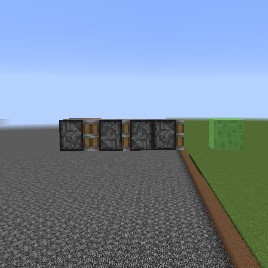}} \hspace{0.01\textwidth}
  \subfloat[FlyingMachine -- step 30]{\label{ref_label3}\includegraphics[width=0.15\textwidth]{ 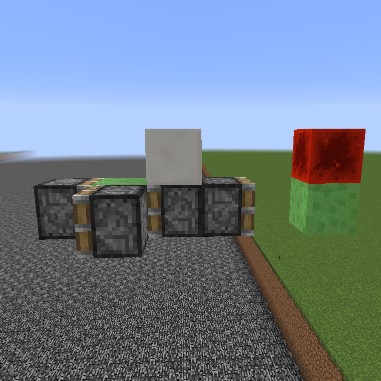}} \hspace{0.01\textwidth}
  \subfloat[FlyingMachine moving -- step 60]{\label{ref_label4}\includegraphics[width=0.15\textwidth]{ 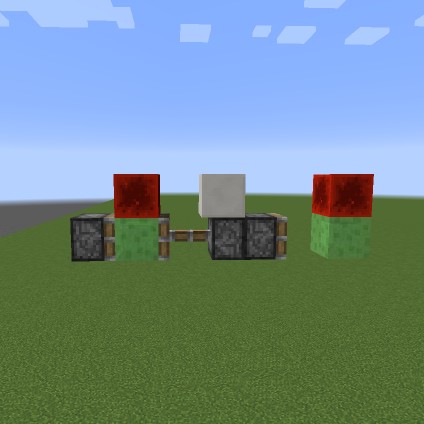}} \hspace{0.01\textwidth}

  \subfloat[Target Caterpillar]{\label{ref_label1}\includegraphics[width=0.15\textwidth]{ 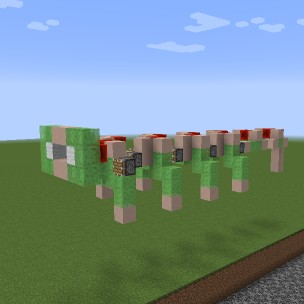}} \hspace{0.05\textwidth}
  \subfloat[Caterpillar -- step 10]{\label{ref_label1}\includegraphics[width=0.15\textwidth]{ 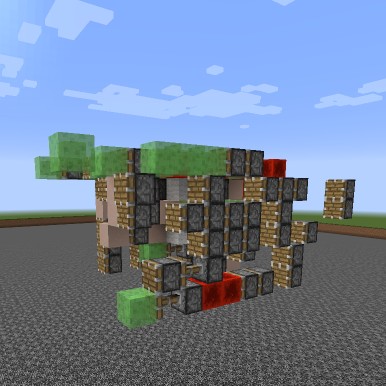}} \hspace{0.01\textwidth}
  \subfloat[Caterpillar -- step 15]{\label{ref_label2}\includegraphics[width=0.15\textwidth]{ 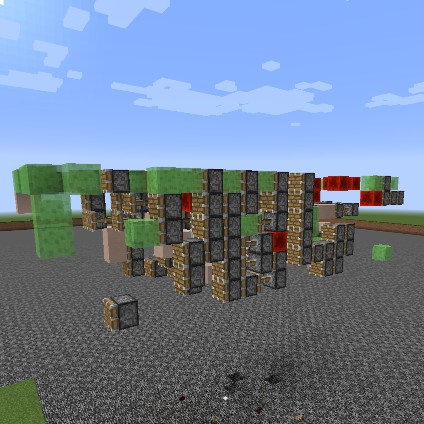}} \hspace{0.01\textwidth}
  \subfloat[Caterpillar -- step 30]{\label{ref_label3}\includegraphics[width=0.15\textwidth]{ 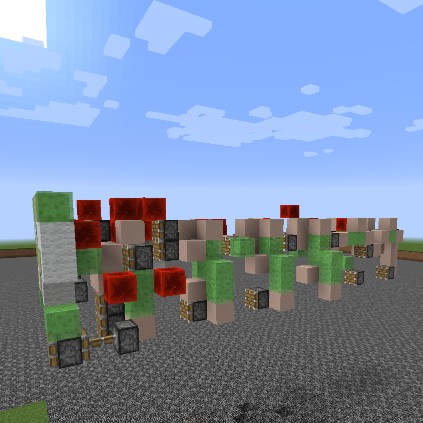}} \hspace{0.01\textwidth}
  \subfloat[Caterpillar moving -- step 60]{\label{ref_label4}\includegraphics[width=0.15\textwidth]{ 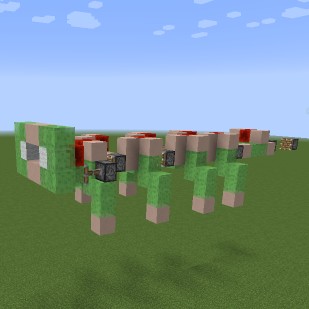}} \hspace{0.01\textwidth}
  \caption{\label{functional_machine_generation} Functional machine generation over time}
\end{figure*}

As an inspiration for the structures we trained the CA to reproduce, we used several publicly available artefacts built by the Minecraft community. Namely we used structures found on \url{www.planetminecraft.com/project/111-structure_block-saves-to-share-nbt} website. This library contains files in the NBT format, which is a JSON-like object that contains block types, positions, etc; everything needed to serialize a MineCraft build for later use. We used EvoCraft API to spawn the NBT files into the running Minecraft server. We added a structure called "apartment complex" from YouTuber Pixlriffs website. Finally, for the moving "redstone" builds we took a caterpillar found in this video \url{www.youtube.com/watch?v=wNqzwAPdFbs} as an inspiration for our caterpillar build (Figure \ref{functional_machine_generation}). 

\subsection{Experimental Details}
Our model is evaluated on a series of regeneration tasks for both static structures and dynamic functional machines, their relevant details are recorded in Table \ref{tab:entity_details}.
We initialize the NCA's convolutional layers using standard normal initialization and zero bias. We run each experiment with a batch size of 5 for 20k steps, with early stoppage when losses are $<$ 0.005. Network size, number of steps, and other relevant hyperparameters are recorded in Table \ref{tab:hyperparam_details}
Some entities, such as functional robots, have redstone and pistons which are orientation sensitive components. Because the NCA model only predicts block type and not orientation, we set these orientations by default to NORTH. 

    Experiments were run using a 2080ti GPU and a Titan RTX GPU. On average, smaller structures took ~1-3 seconds each epoch, and bigger structures took around ~15-24 seconds each epoch. The full library used for training the CA will be made available soon.

\section{Results}

\subsection{Growing static structures}
Despite having simple and local individual updates, our NCA is capable of generating complex structures, some almost identical to the targets. It also scales well to increasing number of individual block types. See Figure \ref{structure_generation} and videos at \url{https://youtu.be/-EzztzKoPeo} for results.

The NCA, contrary to the work presented in \cite{kim2021learn3d}, is able to generate diverse and complex interiors like those of ApartmentBlock and JungleTemple. Surprising to us, in the case of JungleTemple, the NCA even generates a functional arrow trap, which uses a working redstone circuit.
\begin{figure}[H]
  \centering
  \subfloat[ApartmentBlock: tv living room]{\label{ref_label3}\includegraphics[width=0.22\textwidth]{ 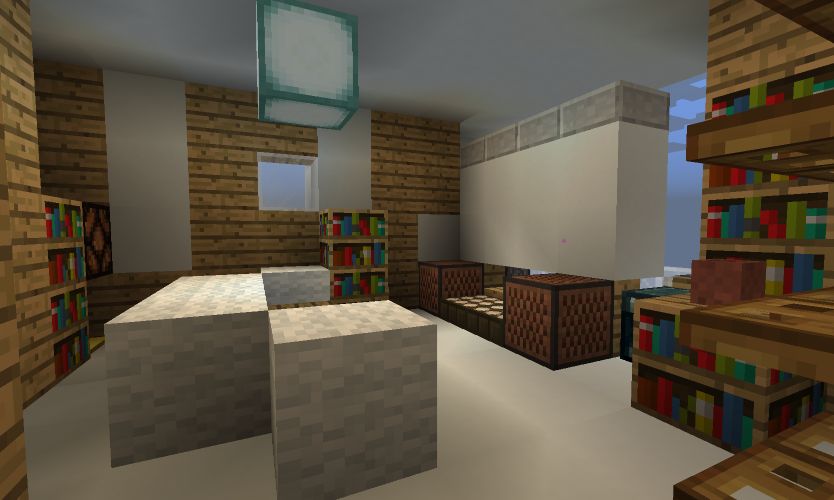}} \hspace{0.01\textwidth}
  \subfloat[JungleTemple: working redstone circuit]{\label{ref_label4}\includegraphics[width=0.22\textwidth]{ 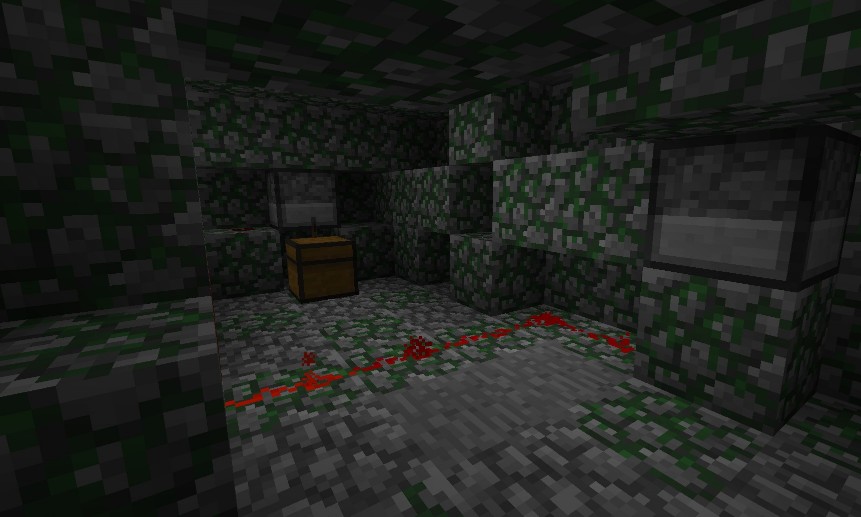}} \hspace{0.01\textwidth}
\end{figure}
The NCA's ability to reconstruct structures is proficient with respect to a wide range of size and number of unique blocks, as seen in the decreasing training curves in Figure~\ref{training_curves}. However, larger entities,  such as the cathedral, are more challenging to reconstruct than their smaller counterparts, as the model is often stuck in local minima and takes much longer to train. Even though the cathedral model achieved a lower loss than some of the other entities, the generated structure has many random artifacts and is less natural than the others (Figure~\ref{cathedral_structure_generation}). Entities that are more random in nature are also harder to generate. This is especially apparent in the oak tree, where the loss over time is very sporadic. However, unlike the cathedral, the trained model generates very natural trees that resemble the target very closely, but this may also be due to the structure's inherent randomness.

\subsection{Growing functional machines}

The NCA was also successful in replicating functional machines (Figure \ref{functional_machine_generation}). As a result, after 60 steps each trained NCA generated a working machine, even a complex one like the caterpillar. Even though all the different NCAs shared the same initial seed, we observed different growth patterns: the flying machine started generating small cells then gradually expanding out into the final functional state. The caterpillar, however, started growing wildly in the earlier steps, then culling cells to form the final functional state.

\begin{figure}[!htb]
  \centering
  \subfloat[Caterpillar Cut in half]{\label{ref_label1}\includegraphics[width=0.19\textwidth]{ 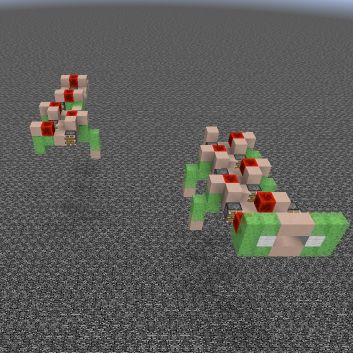}} \hspace{0.01\textwidth}
  \subfloat[Caterpillars regenerating -- step 12]{\label{ref_label2}\includegraphics[width=0.19\textwidth]{ 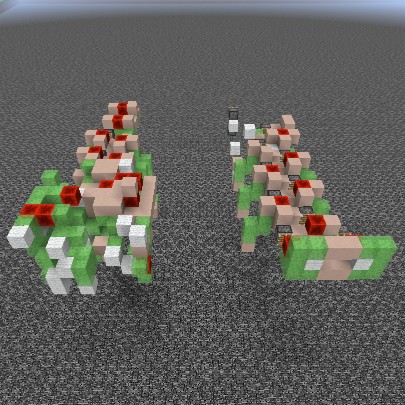}} \hspace{0.01\textwidth}
  \subfloat[Caterpillars regenerating -- step 24]{\label{ref_label3}\includegraphics[width=0.19\textwidth]{ 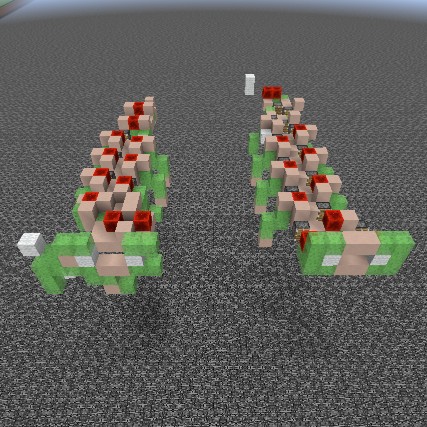}} \hspace{0.01\textwidth}
  \subfloat[Caterpillars regenerated \& moving]{\label{ref_label4}\includegraphics[width=0.19\textwidth]{ 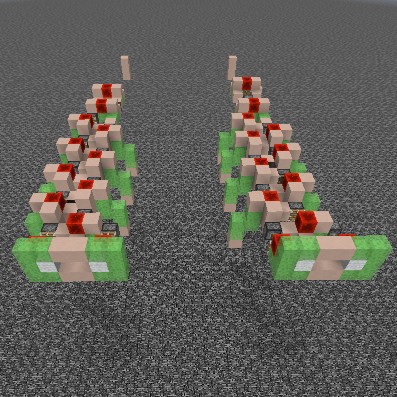}} \hspace{0.01\textwidth}
 \caption{\label{catepillar_regeneration} Caterpillar's regenerating after being sliced in half.}
\end{figure}

\begin{figure}[!htb]
  \centering
  \subfloat[Tree cut in half]{\label{ref_label1}\includegraphics[width=0.19\textwidth]{ 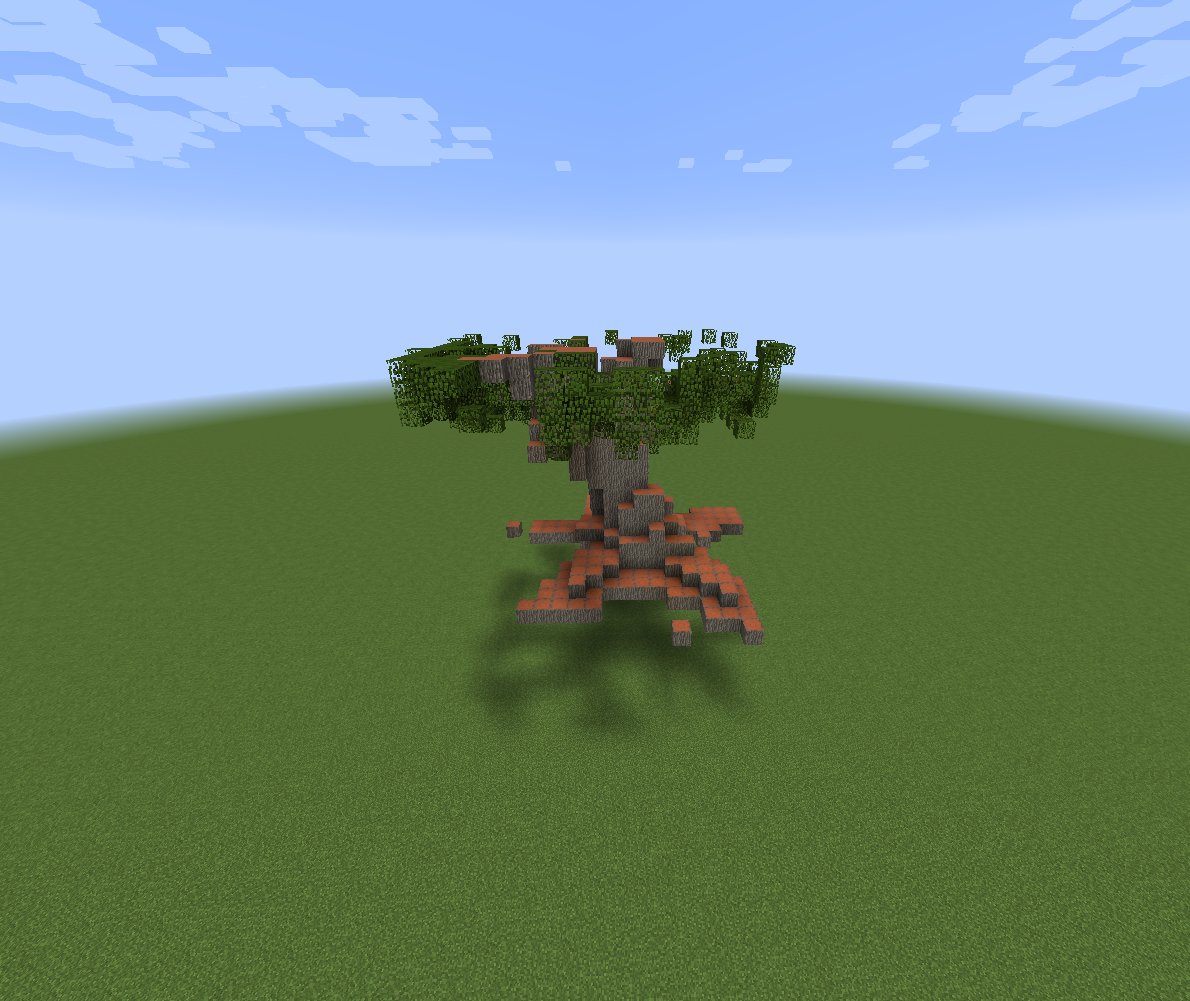}} \hspace{0.01\textwidth}
  \subfloat[Tree - step 96 
  ]{\label{ref_label4}\includegraphics[width=0.19\textwidth]{ 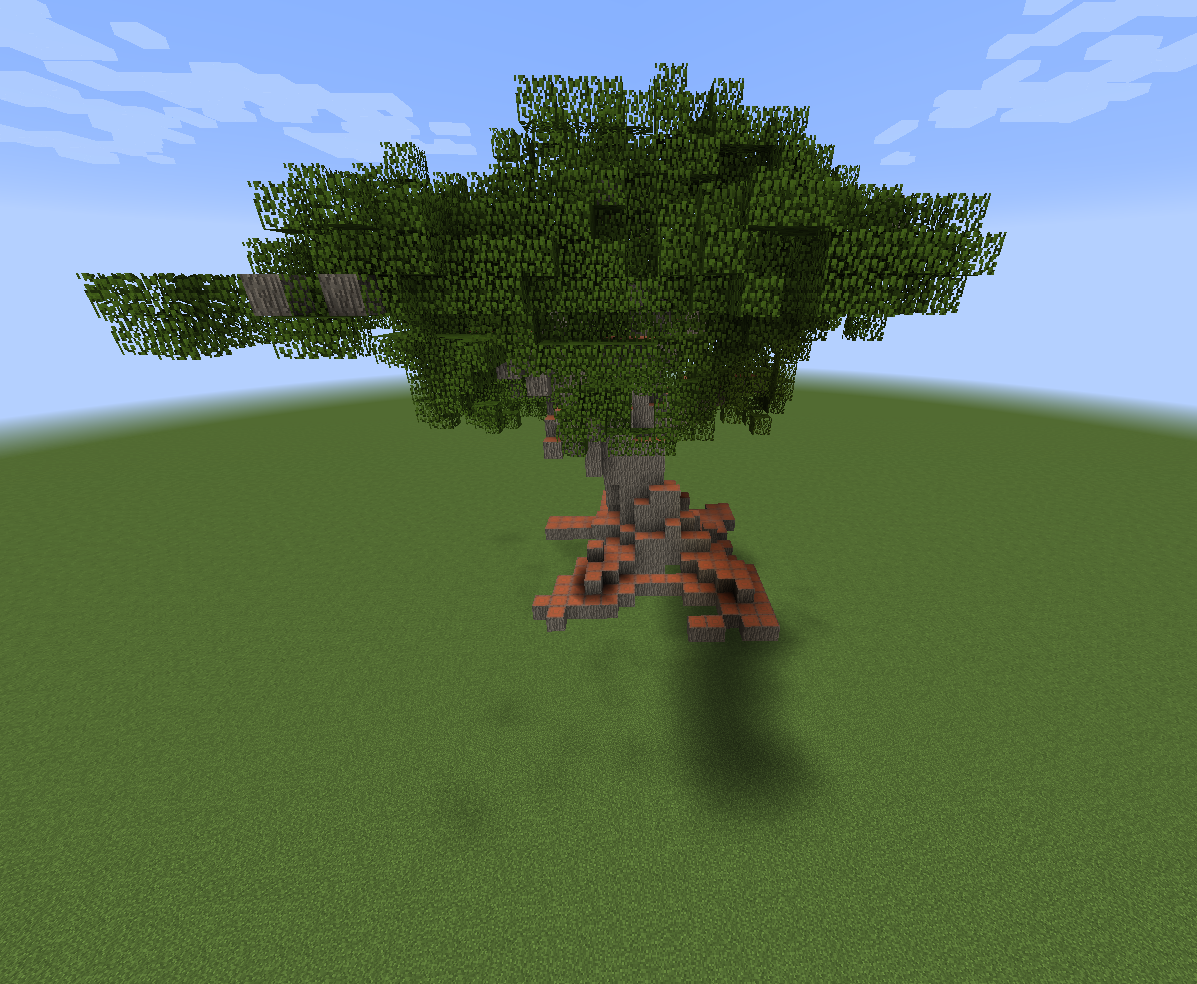}} \hspace{0.01\textwidth}
 \caption{\label{tree_regeneration} Tree regenerating after cut in half, without training to be regenerated.}
\end{figure}

\subsection{Regenerative Properties}

One of the attractive properties of NCAs are their similarity to the natural process of developing complex organisms from a single cell only using few parameters encoding local update rules. Beside growing complex structures from a single cell, these local update rules also allow for regenerating or repairing damage. NCAs were shown to be able to regrow parts of their structure when damaged, even when they are not specifically trained for regeneration \citep{mordvintsev2020growing}. This ability can be enhanced further by specifically training the model to be able to recover from damaged states, by providing damaged variants of the samples from the sample pool during training.

The metric we used for evaluating regeneration capability is the ratio of regenerated blocks after the structure was cut in half. We compare a model trained without regeneration and a model trained with regeneration on the task of regenerating a Caterpillar (functional machine). When the model was trained for regeneration, it was able to regrow 99\% of the blocks in case of the Caterpillar when cut in half (Figure~\ref{catepillar_regeneration}). When the model was not trained for regeneration, it only regenerated 30\% of the Caterpillar. 


Even when the NCA is not trained for regeneration, it still has some ability to regenerate from damage for certain structures such as the tree (Figure~\ref{tree_regeneration}), while less so on functional machines. However it is clear that a model trained for regeneration performs much better, while also being able to grow entities at the same level as its counterpart. 

\section{Discussion and Future Work}
In this paper, we propose an improved Neural Cellular Automata from \cite{mordvintsev2020growing} that reproduces 3D Minecraft structures and functional robots with high accuracy. We modify the original automata to include 3D convolutional layers with learnable parameters, cell state vector representations compatibile with the Minecraft voxel environment, and a combined loss function with an emphasis on structural resemblance to the target. To assess our model's efficacy, we teach our NCA to generate a set of Minecraft structures, including both static architectures and simple functional robots. The NCA is able to generate structures that are, in some cases, identical to their respective targets. In addition, these generations are robust to perturbations and damage, able to recover blocks that are destroyed.

Although the NCA demonstrates promising results in complex 3D structure regeneration, there is still room for improvement. Our approach is not robust enough, since the model has to be trained separately for every new entity, making the model parameters unsharable between tasks. This may be relieved by adding a global one-hot vector for all possible block types. Such an improved embedding can be extended multiple entities. Another approach would be to adopt the autoencoder from \cite{ruiz2021neural} to encode entities' structural information.

Another potential improvement would be to shift from supervised to reinforcement learning. Reinforcement learning have demonstrated potency in complex environments \citep{lillicrap2015continuous,vinyals2019grandmaster,mnih2013playing}, even with sparse rewards \cite{vecerik2017leveraging}. We can take advantage of SOTA reinforcement learning algorithms to augment the iterative training process of the NCA.

Training only on reconstruction tasks would not teach the NCA the inner workings for a machine to be functional. A potential approach to facilitate understanding of mechanisms is via additional contextual costs (e.g. evaluating a reconstruction of a flying machine by the distance travelled by the generation). This would allow for a more diverse set of generated machines.

We also look forward to attempting to generate increasingly complex and larger machines, such as a functional redstone computer. This would require the model to become much more efficient and utilize multi-gpu training procedures such as  \cite{rajbhandari2020zero}

\section{Conclusion}

Our paper broadens the scope of modalities for tasks suitable for Neural Cellular Automata. Although the work presented is in a simplified 3D environment, the model's capability of generating increasingly complex 3D entities brings us one step closer to real-life, self-organizing, and regenerative physical artefacts. Self-organizing capabilities in 3D can lead to more scalable, even growable, physical systems. 
Furthermore, regenerative abilities in NCA may inspire self-repairing structures, including buildings and artificial organs. Through investigating and expanding the capabilities of Neural Cellular Automata, we hope to facilitate further developments in various fields of engineering.


\section{Acknowledgments}

This project was supported by a Sapere Aude: DFF-Starting Grant (9063-00046B) and compute through  an Amazon Research Award.
\footnotesize
\bibliographystyle{apalike}
\bibliography{main} 

\end{document}